\documentclass[sigconf]{aamas}
\usepackage{balance} % for balancing columns on the final page
\usepackage[capitalise]{cleveref}

\setcopyright{ifaamas}
\acmConference[AAMAS '22]{Proc.\@ of the 21st International Conference
on Autonomous Agents and Multiagent Systems (AAMAS 2022)}{May 9--13, 2022}
{Online}{P.~Faliszewski, V.~Mascardi, C.~Pelachaud,
M.E.~Taylor (eds.)}
\copyrightyear{2022}
\acmYear{2022}
\acmDOI{}
\acmPrice{}
\acmISBN{}

\acmSubmissionID{???}

\title[SIERRA]{SIERRA: A Modular Framework for Research Automation }
\subtitle{Demonstration Track}

\author{John Harwell}
\affiliation{
  \institution{University of Minnesota}
  \country{United States}}
\email{harwe006@umn.edu}

\author{London Lowmanstone}
\affiliation{
  \institution{University of Minnesota}
  \country{United States}}
\email{lowma016@umn.edu}

\author{Maria Gini}
\affiliation{
  \institution{University of Minnesota}
  \country{United States}}
\email{gini@umn.edu}

\begin{abstract}
  Modern intelligent systems researchers employ the scientific method: they form
  hypotheses about system behavior, and then run experiments using one or more
  independent variables to test their hypotheses.  We present SIERRA, a novel
  framework structured around that idea for accelerating research developments
  and improving reproducibility of results. SIERRA makes it easy to quickly
  specify the independent variable(s) for an experiment, generate experimental
  inputs, automatically run the experiment, and process the results to generate
  deliverables such as graphs and videos. SIERRA provides reproducible
  automation independent of the execution environment (HPC hardware, real
  robots, etc.) and targeted platform (arbitrary simulator or real robots),
  enabling exact experiment replication (up to the limit of the execution
  environment and platform). It employs a deeply modular approach that allows
  easy customization and extension of automation for the needs of individual
  researchers, thereby eliminating manual experiment configuration and result
  processing via throw-away scripts.
\end{abstract}


\newcommand{\BibTeX}{\rm B\kern-.05em{\sc i\kern-.025em b}\kern-.08em\TeX}

\begin{document}

%%% The following commands remove the headers in your paper. For final
%%% papers, these will be inserted during the pagination process.

\pagestyle{fancy}
\fancyhead{}

%%% The next command prints the information defined in the preamble.

\maketitle

%%%%%%%%%%%%%%%%%%%%%%%%%%%%%%%%%%%%%%%%%%%%%%%%%%%%%%%%%%%%%%%%%%%%%%%%

\section{Introduction}

In modern intelligent agents research, reproducibility is crucial. Configuring
experiments for testing theories can be time consuming, and further compounded
by the need to deal with different configurations for specific platforms (e.g.,
ROS~\cite{ROS2009}) and execution environments (e.g., SLURM~\cite{SLURM2003}
clusters).  Ad hoc tools and scripts are frequently developed to meet these
needs on a per-project basis, and reused or modified as needed between projects.
Scripts for processing and visualizing experimental results are developed
similarly.  In this paper, we present
SIERRA\footnote{\url{https://github.com/swarm-robotics/sierra.git}}, a framework
to automate the process of hypothesis testing and results processing. SIERRA is
complementary to other reproducibility-enhancing tools for AI research such as
Robotarium~\cite{Robotarium2016} and IEEE Code Ocean.

SIERRA is \emph{modular}, and can be extensively customized to meet a diverse
range of researcher needs. Originally developed for robotics research, its
modular design makes it capable of supporting almost any platform and execution
environment in AI research.

SIERRA experiments are fully \emph{reproducible}, regardless of the platform
targeted and the execution environment on which the experiments are run. SIERRA
currently supports the ARGoS~\cite{Pinciroli2012}, Gazebo~\cite{Gazebo2004}, and
ROS~\cite{ROS2009} platforms. To the best of our knowledge no automation exists
for these platforms for hypothesis testing and results processing---crucial
parts of intelligent systems research. Some partial automation of
Webots~\cite{Webots2004} was done in~\cite{WebotsHPC2021}; such automation is a
subset of SIERRA's capabilities. Independent from the targeted platform, SIERRA
currently supports several execution environments for running experiments:
PBS/SLURM clusters, real robots, and the local machine. New execution
environment or platform support can be easily added as python modules. Through
automation, SIERRA streamlines the process of publishing reproducible research:
all automation provided by SIERRA is idempotent.

\begin{figure*}[t]
  \centering
  \includegraphics[width=.95\linewidth,height=7cm]{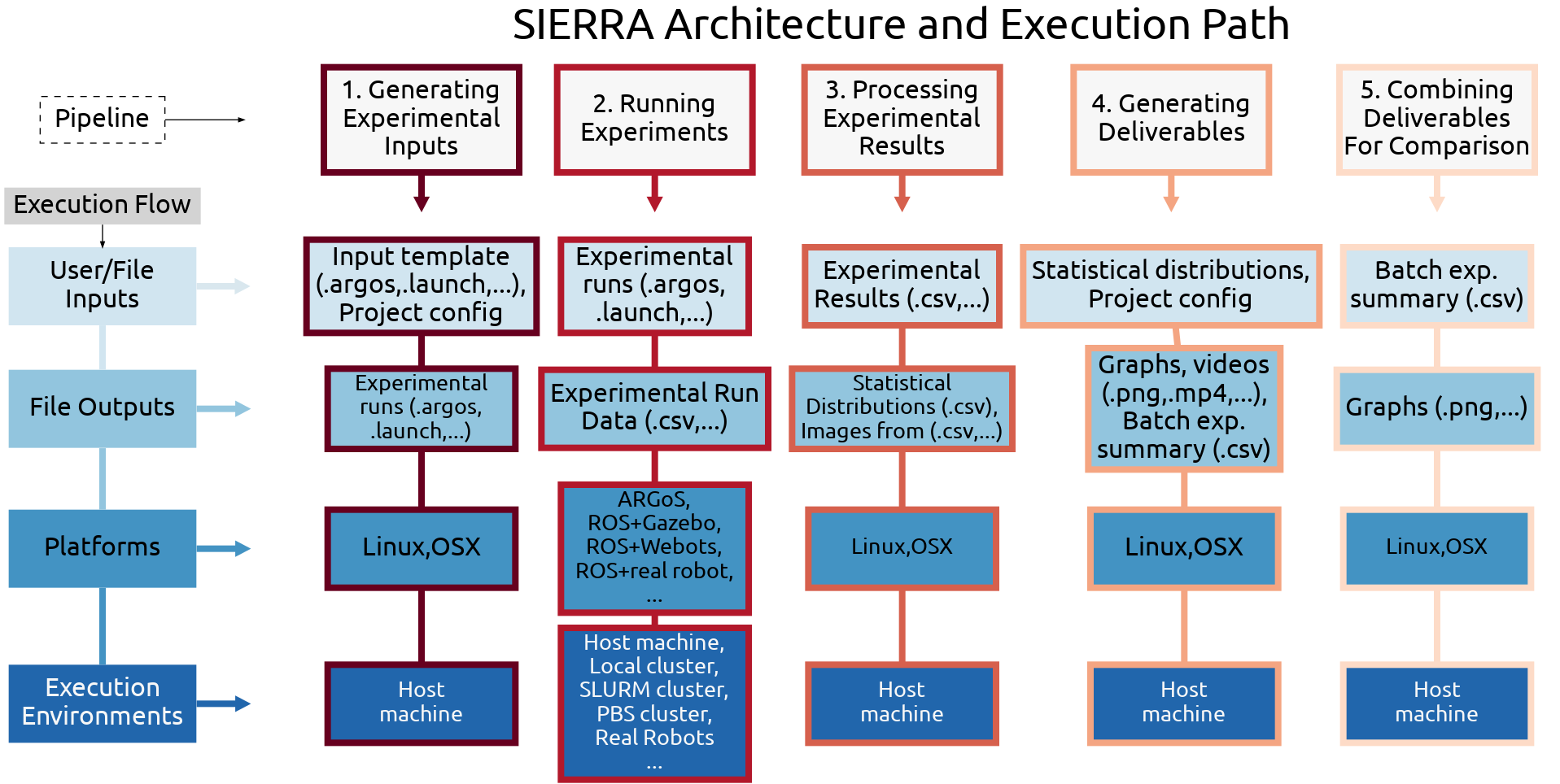}
  \caption{\label{fig:arch} \footnotesize{Architecture of SIERRA, organized by
      pipeline stage. Pipeline stages are listed left to right, with an
      approximate joint architectural/functional stack from top to bottom for each
      stage. ``\dots'' indicates areas where SIERRA is designed via python
      plugins to be easily extensible. ``Host machine'' indicates the machine
      SIERRA was invoked on.}}
\end{figure*}

SIERRA is a \emph{cohesive framework} for end-to-end automation of the
scientific method in intelligent systems research. SIERRA automates the pipeline
described in~\cref{sec:overview} and~\cref{fig:arch}, eliminating the need for
throw-away scripts, greatly reduces menial re-configuration of experimental
inputs across platforms and execution environments, and provides a uniform
interface for generating camera-ready deliverables such as graphs and videos for
inclusion in academic papers. Essentially, SIERRA handles the ``backend'' parts
of research, allowing researchers to focus on the ``research'' aspects:
developing theories and testing hypotheses via experimental evaluation.

\section{SIERRA Overview}\label{sec:overview}

SIERRA supports \emph{research queries}: a hypothesis expressed as a query of an
independent variable over some range, expressed in a user-defined command line
syntax (e.g., \verb|population_size.Log8| for representing varying the number of
agents across \{1,2,4,8\}). The expressed query is used to define a \emph{batch
  experiment}: a set of \emph{experiments} each with a different value of the
independent variable (number of agents in the above example). Each experiment
consists of one or more \emph{experimental runs} which can be simulation runs or
real robot trials. For some types of research, multiple experimental runs are
required due to randomness (e.g., imperfect sensors/actuators on real robots);
SIERRA provides configurations for managing this complexity and for generating
and plotting statistical distributions over results, and handles all necessary
platform and execution environment configuration. However, it does not attempt
to homogenize configurations such that the results of a research query can be
answered in a platform and execution environment independent way (which is not
possible in general).

SIERRA has been used for several publications in top
conferences~\cite{Harwell2020a,Harwell2019a} using ARGoS and PBS/SLURM HPC
clusters. In this work, we further demonstrate its utility by generating
experiments for ROS and real
robots~\footnote{\url{https://www-users.cse.umn.edu/~harwe006/showcase/aamas-2022-demo}}. We
also demonstrate how SIERRA can be used to explore independent variables in a
short
demonstration~\footnote{\url{https://www-users.cse.umn.edu/~harwe006/showcase/aamas-2022-demo/\#demonstration-experimental-results-visualization-stage-4}}.
SIERRA addresses the need for reproducibility and research acceleration in
intelligent systems, and we therefore strongly argue for its inclusion in any
researcher's toolbox.

% [JRH] Background stuff will probably have to be here too...

\subsection{Experimental Input Generation}
%
% [JRH] This would be nice to include, but cut for space.
%% SIERRA currently requires that the template input file is XML, which was
%% chosen over other input formats because (1) it is not dependent on
%% whitespace/tab/spaces for correctness, making it more robust to multiple
%% platforms, simulators, parsers, users, etc., and (2) mature manipulation
%% libraries exist for python and C++, two of the most population intelligent
%% system development languages, so it should be relatively straightforward for
%% projects to read experimental definitions from XML. Many platforms already
%% support XML input (e.g., ARGoS, ROS, WeBots), and if a research wants to add
%% SIERRA support for a platform which does not support XML, SIERRA's modular
%% architecture makes it easy to do so.
%
To generate the batch experiment, a template XML file is modified according to
the research query, with one experiment generated for each ``value'' of the
independent variable(s), so that by comparing results across experiments in the
batch, changes in behavior in response to the different ``values'' can be
observed. Each ``value'' may correspond to a single change to the template via
(e.g., \verb|population_size.Log8| for changing the number of agents), or it can
correspond to multiple changes (e.g., testing across a set of environments which
require setting multiple XML parameters for each one). SIERRA also supports
changing additional parts of the template input file uniquely for each
experiment in the batch, or uniformly for all experiments, providing
unparalleled expressiveness to support research automation through experiment
generation.

\subsection{Running Experiments}
Experimental run inputs are executed on some simulator or real-robot
\emph{platform} in some \emph{execution environment}. For example a researcher
may want to run simulations targeting ARGoS~\cite{Pinciroli2012} on a SLURM
managed cluster, or to run experiments generated from the \emph{same} research
query on the ROS~\cite{ROS2009} platform with a real robot (e.g.,
Turtlebot3~\cite{Turtlebot2020}). Switching between platforms and/or execution
environments naturally requires some code changes (e.g.,
cross-compiling). However, a lot of additional researcher time is spent figuring
out how to get an experiment to run within a given environment, which slows down
the actual research. SIERRA handles this complexity, reducing the burden on
researchers.

% However, given the same research query, modifying experimental
% inputs according to changes between platforms or execution environments imposes
% a large burden on researchers unrelated to the research itself.
\subsection{Processing Experimental Results}
Outputs from executed experiments are processed. This includes statistical
distribution generation across experimental runs for each experiment in a batch,
as well as across experiments in a batch, and converting output .csv files into
images which can be stitched together into videos during stage 4.

\subsection{Generating Deliverables}
Processed experimental results are used to generate deliverables to accompany or
be part of published research. This can include graphs or videos showing
different aspects of the system's response to the research query. SIERRA's
automation in this stage makes it easy to modify a specific graph or video, if,
for example, one needs to be modified at a reviewer's request, eliminating the
tedious process of locating previously written throw-away scripts to regenerate
it, again saving time. Furthermore, this automation also improves
reproducibility: with a single SIERRA command on a properly configured
environment researcher B could reproduce the exact results of researcher A if
both were using SIERRA and they had access to researcher A's agent code.

\subsection{Deliverable Comparison}
After deliverable generation, multiple deliverables are combined to provide
side-by-side comparisons. For example, comparing the performance of a new
reinforcement learning method against the current state-of-the-art method, each
of which was independently generated during the previous stage may be
desirable.

%%%%%%%%%%%%%%%%%%%%%%%%%%%%%%%%%%%%%%%%%%%%%%%%%%%%%%%%%%%%%%%%%%%%%%%%

%%% The acknowledgments section is defined using the "acks" environment
%%% (rather than an unnumbered section). The use of this environment
%%% ensures the proper identification of the section in the article
%%% metadata as well as the consistent spelling of the heading.

%% \begin{acks}

%% \end{acks}

%%%%%%%%%%%%%%%%%%%%%%%%%%%%%%%%%%%%%%%%%%%%%%%%%%%%%%%%%%%%%%%%%%%%%%%%

%%% The next two lines define, first, the bibliography style to be
%%% applied, and, second, the bibliography file to be used.
\newpage
\bibliographystyle{ACM-Reference-Format}
\bibliography{2022-phd-thesis,2022-aamas-demo}

%%%%%%%%%%%%%%%%%%%%%%%%%%%%%%%%%%%%%%%%%%%%%%%%%%%%%%%%%%%%%%%%%%%%%%%%

\end{document}